\documentclass[10pt,twocolumn,letterpaper]{article}

\usepackage{iccv}
\usepackage{times}
\usepackage{graphicx}
\usepackage{amsmath}
\usepackage{amssymb}

\usepackage{booktabs}
\usepackage{algorithm}
\usepackage{algpseudocode}
\newtheorem{theorem}{Theorem}
\newtheorem{corollary}{Corollary}[theorem]

\usepackage{multirow}
\usepackage[table]{xcolor}
\usepackage{caption}
\usepackage{subcaption}
\usepackage{tablefootnote}

\usepackage[pagebackref=true,breaklinks=true,letterpaper=true,colorlinks,bookmarks=false]{hyperref}

\usepackage[breaklinks=true,bookmarks=false]{hyperref}

\iccvfinalcopy 

\def\iccvPaperID{****} 

\ificcvfinal\fi

\begin{document}

\title{DPPMask: Masked Image Modeling with Determinantal Point Processes}

\author{
Junde Xu \textsuperscript{1, 2}
\thanks{Equal Contribution.} \and
Zikai Lin \textsuperscript{1, 2 *}  \and
Donghao Zhou \textsuperscript{1, 2} \and
Yaodong Yang \textsuperscript{4} \and
Xiangyun Liao \textsuperscript{1} \and 
Bian Wu \textsuperscript{3} \and
Guangyong Chen \textsuperscript{3}\thanks{Corresponding Author (\href{mailto:gychen@zhejianglab.com}{gychen@zhejianglab.com}).} \and
Pheng-Ann Heng \textsuperscript{1, 4} \and
\textsuperscript{1} Shenzhen Institute of Advanced Technology, Chinese Academy of Sciences \and
\textsuperscript{2} University of Chinese Academy of Sciences \and
\textsuperscript{3} Zhejiang Lab \and
\textsuperscript{4} The Chinese University of Hong Kong
       }

\maketitle
\ificcvfinal\thispagestyle{empty}\fi


\begin{abstract}
    Masked Image Modeling (MIM) has achieved impressive representative performance with the aim of reconstructing randomly masked images.
    Despite the empirical success, most previous works have neglected the important fact that it is unreasonable to force the model to reconstruct something beyond recovery, such as those masked objects.
    In this work, we show that uniformly random masking widely used in previous works unavoidably loses some key objects and changes original semantic information, resulting in a misalignment problem and hurting the representative learning eventually.
    To address this issue, we augment MIM with a new masking strategy namely the DPPMask by substituting the random process with Determinantal Point Process (DPPs) to reduce the semantic change of the image after masking.
    Our method is simple yet effective and requires no extra learnable parameters when implemented within various frameworks.
    In particular, we evaluate our method on two representative MIM frameworks, MAE and iBOT.
    We show that DPPMask surpassed random sampling under both lower and higher masking ratios, indicating that DPPMask makes the reconstruction task more reasonable.
    We further test our method on the background challenge and multi-class classification tasks, showing that our method is more robust at various tasks.
\end{abstract}

\section{Introduction}
\label{sec:intro}

Self-supervised learning aims to extract semantic features by solving auxiliary prediction tasks (or pretext tasks) with pseudo labels generated solely based on input features.
While various tasks have been proposed for self-supervised learning, one intuitive idea is learning representations by recovering the original data from the corrupted structure.
The philosophy behind such methods is simple: what the model generates can examine whether the model understands.
This principle was first introduced by the denoising autoencoder  \cite{vincent2010stacked} which has supported significant advances in NLP \cite{brown2020language, radford2019language, liu2019roberta}.
Methods that follow this idea such as BERT \cite{devlin2018bert} now have become a dominant routine.
In the field of image processing tasks, although reconstruction-based pre-training was first put forth by  \cite{pathak2016context}, it wasn't until recently that methods based on this concept were brought back to state-of-the-art performance.
Benefiting from the new network architectures like ViT \cite{dosovitskiy2020image}, Masked Image Modeling (MIM) has become highly popular, and there is a series of more aggressive masking strategies like MAE \cite{he2022masked}, simMIM \cite{xie2022simmim}.
\begin{figure}[t]
  \centering
    \includegraphics[width=0.9\linewidth]{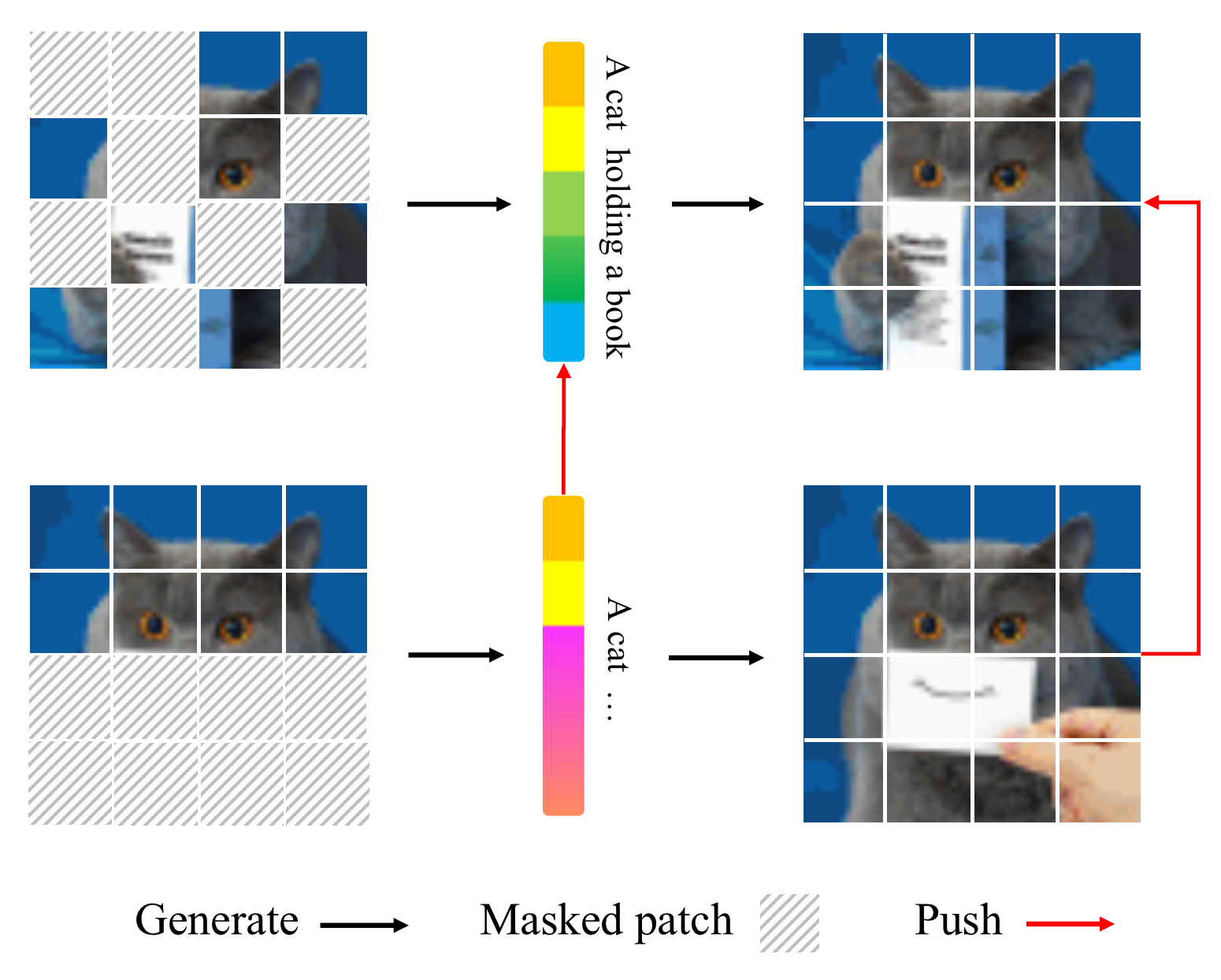}
  \caption{Illustration of the misalignment problem in MIM. The model generates predictions that differ plausibly from the original image, while the original image has still been imposed as supervision, leading to an unreasonable high loss.}
  \label{fig:1}
  \vspace{-3mm}
\end{figure}  

However, simply random masking can be problematic in practice. 
An important fact is that it is unreasonable to force the model to reconstruct something beyond recovery.
Consider a simple case as Fig.~\ref{fig:1}.
The top row of Fig.~\ref{fig:1} shows the underlying logic of MIM: successful reconstruction implies the network captures the correct semantic features.
While the bottom row of Fig.~\ref{fig:1} shows a failure case: if the masking process happens to drop an important semantic of the original image, the book in Fig.~\ref{fig:1}, then will change the semantics of the original image and make the network hardly recover it from the rest.
In this case, if the model continues to be forced to reconstruct the original image, the model might fill in the obscured image with whatever is feasible, which will interfere with the process of learning the original features.
Furthermore, as the masking rate increases, the original semantic information is distorted with a higher probability.
We refer to such a situation as a misalignment problem, \textit{i.e.} the semantics of masked image and the original image are miss-aligned.
Consequently, the misalignment problem will cause the alignment of improper sample pairs, which will eventually harm the performance of the downstream task.

Some studies also proposed constructing better sample pairs for MIM.
In MaskFeat \cite{wei2022masked}, they change the pixel reconstruction task to HOG reconstruction, to reduce the impact of some ambiguous situations for the network to prediction, such as colors, and textures.
However, MaskFeat can only reduce the impact of hardly recoverable high-frequency signals. 
If the masked part contains the whole object instance, there is still not enough information for the network to rebuild.
In ADIOS \cite{shi2022adversarial} and SemMAE \cite{li2022semmae}, they train an extra segmentation network to partition the image into different semantics. 
However, the number of semantics varies significantly in different images, thus, it is hard to find an optimal semantic partition network. 
In AttMask \cite{kakogeorgiou2022hide}, they mask the most attended patches according to their attention score to construct more challenging MIM tasks.
However, AttMask needs an attention map to perform sampling, which can not fit into
reconstruction-based MIM methods (e.g., MAE) seamlessly.
Recently, AMT \cite{gui2022good} adding attention map guided masking to MAE.
Unfortunately, these algorithms do not take into account the misalignment problem, which leads to inferior performance.

In contrastive learning, an \textit{InfoMin} principle suggests that two augmented views of an image should retain task-relevant information while minimizing irrelevant nuisances \cite{tian2020makes}.
Analog to MIM, we can summarize the following two conditions:
First, the selected patches should be representative enough to cover the whole semantic information of the original image.
Second, the masking ratio should be set to a high level to minimize the irrelevant information shares between different masks of the same image.
While the second constraint is easy to satisfy, the problem is how to retain the task-relevant original semantics under the limited input ratio. 
To address this, we propose a novel masking strategy based on Determinantal Point Process (DPPs).
DPPs are elegant probabilistic models on sets that can capture both quality and diversity when a subset is sampled from a ground set \cite{kulesza2012determinantal, launay2021determinantal}, making them ideal for modeling the set that contains more information of original images as possible.
During the sampling process, DPPs will compute the distance of each patch, and select patches that are dis-similar from the selected subset.
This process makes the network focus on the patches with more representative information.
For example, the unique color, texture, etc. 
We show that our new sampling strategy can obtain more representative patches to keep the semantics unchanged and alleviate the impact of the misalignment problem.
More importantly, We show that DPPMask surpassed random sampling under both lower and higher masking ratios, indicating that DPPMask makes the reconstruction task more reasonable.   
Furthermore, our method needs no extra training process and achieves minimal computational resource consumption. 

Our contribution can be summarized as follows:
\begin{itemize}
    \item We analyze the training behavior of reconstruction-based MIM and discuss the impact of the misalignment problem.
    \item To alleviate the impact of misalignment in MIM, we proposed a novel plug-and-play sampling strategy called DPPMask based on DPPs. Our method can generate more reasonable training pairs, is simple yet effective, and requires no extra learning parameters.
    \item We verify our method on two representative MIM frameworks, our experiments evidence that features learned by fewer misalignment problems achieve better performance in downstream tasks.
\end{itemize}
\begin{figure*}[t]
  \centering
  \includegraphics[width=1.0\linewidth]{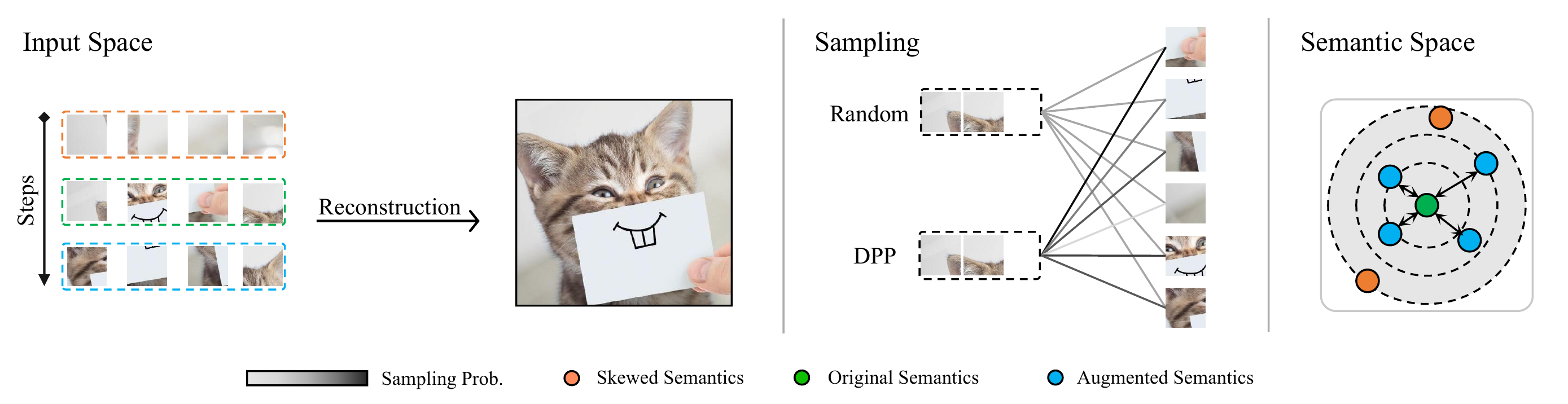}
  \caption{An illustration of misalignment and our method. For each image, the masking policy can propose different patch sets for reconstruction, however, some of them with skewed semantics are not suitable for reconstruction. Our method models the probability of co-occurrence of each patch to avoid such misalignment problems.}
  \label{fig:MIM}
  \vspace{-3mm}
\end{figure*}

\section{Related Work}
\label{sec:rw}
\noindent\textbf{Self-supervised learning.} Classic deep learning trains the parameters of the model by utilizing labeled data. 
Instead, self-supervised learning(SSL) expects to acquire representations with unlabeled data by a pre-text task.
Among them, Masked language modeling (MLM) has taken the lead to be a highly influential self-supervised learning model before.
$e.g.,$, BERT  \cite{devlin2018bert} and GPT  \cite{radford2018improving,radford2019language} are such successful methods that the academia has focused on these two models for pre-training in NLP.
These models leverage visible tokens in a sequence and predict invisible tokens to gain appropriate representations, which have been proved to successfully repaint the field  \cite{brown2020language}.
In other fields of SSL, there have been numerous methods that focus on different pretext tasks like reconstructing original tokens from image/patch operations  \cite{caron2018deep,doersch2015unsupervised,gidaris2018unsupervised,noroozi2016unsupervised,wei2019iterative}and Spatio-temporal operation  \cite{pathak2017learning,misra2016shuffle,goroshin2015unsupervised,fernando2017self,wang2017transitive}.
A well-known method is contrastive learning that capitalizes on augmentation invariance in the feature space and could be evaluated by linear probing \cite{caron2020unsupervised,caron2021emerging,chen2020simple, chen2021exploring,feichtenhofer2021large,he2020momentum,qian2021spatiotemporal}, which was the previous mainstream based on SSL.
\vspace{0.1cm}

\noindent\textbf{Masked Image Modeling(MIM)} Masked Image Modeling recently has shown capability to reconstruct pixels  \cite{atito2021sit} from corrupted images. 
MIM can be seen as a generalized Denoising AutoEncoders (DAE) \cite{vincent2008extracting,chen2020rewon}, 
which aims to reconstruct the original tokens from corrupt input.
e.g., inputing missing color channels \cite{zhang2016colorful} or missing pixels  \cite{vincent2010stacked}.
Context Encoder  \cite{pathak2016context} reconstructs a rectangle area of the original images using convolutional networks.
Then ViT  \cite{dosovitskiy2020image} and iGPT  \cite{chen2020generative}
recall the learning approach of predicting patches with a contrastive predictive coding loss on the modern vision Transformers, and show strong potential in representation learning.
BEiT  \cite{bao2021beit} proposes to use a pre-trained discrete VAE  \cite{ramesh2021zero}as the tokenizer, and improves MIM's performance further.
However, the tokenizer needs to be offline pre-trained with matched model and dataset which limits its adaptivity.
To end this, iBOT  \cite{zhou2021ibot} presents a new framework that performs masked prediction with an online tokenizer and gains prominence achievement.
Recently, equipped with a more aggressive masking strategy, SimMIM  \cite{xie2022simmim} and MAE  \cite{he2022masked} further demonstrate that simple pixel reconstruction can achieve competitive results from previous pre-training methods.
\vspace{0.1cm}
 
\noindent\textbf{Determinantal Point Process.}
Determinantal point processes (DPPs) are probabilistic models of configurations that favor diversity  \cite{macchi1975coincidence}.
Its repulsion brings new potential to enhance diversity in multiple machine
learning problems, such as feature extract from high dimensional data  \cite{belhadji2020determinantal},
texture synthesis in image processing  \cite{launay2019determinantal},
building informative summaries by selecting diverse sentences  \cite{kulesza2012determinantal}.

In addition to DPPs, there are some previous methods like Markov random fields(MRFs).
However, MRF assumes that repulsion does not depend on context too much, so it cannot express that, say, there can be only a certain number of selected items overall  \cite{kulesza2012determinantal}.
The DPPs can naturally implement this kind of restriction through the rank of the kernel.

\section{Methods}
In this section, we first discuss the impact of the misalignment problem on the downstream tasks, then, we give a brief introduction to DPPs.
Finally, we propose our method of applying DPPs in the patch masking process.
\subsection{Misalignment in MIM}
\label{sec:ci}
Let $x_1, x_2, z$ be the masked image, reconstruction target, and hidden vector.
In the ideal situation, the input $x_1$ and reconstruction target $x_2$ all followed an identical distribution $x_1, x_2 \sim P(z)$.
However, consider the input only captures the partial information of the original image, in this case, $x_1 \sim P(z')$ where $z'$ defines a different distribution to $z$.
For pixel reconstruction tasks, we denote the encoder and the decoder parameterized by $\phi$ and $\theta$ respectively.
Following  \cite{bao2021beit}, MIM training can be viewed as variational autoencoder training  \cite{kingma2013auto}, which can be described as a Maximum A Posteriori (MAP) Estimation: 
\begin{equation}
    \label{eq:target_collapse}
    \mathop{\arg\max}\limits_{\phi, \theta} \mathbb{E}_{z'\sim q_{\phi}(z'|x_1)} \log{ p_\theta (x_2|z')}
\end{equation}
Suppose the encoder is capable of capturing the semantic information of $x_1$ and the decoder is capable of recovering the image described by $z$ and $z'$, by unfolding the decoder part, we get:
\begin{equation}
    \label{eq:target_unfold}
    p_\theta(x_2|z') = \int p_\theta(x_2|z)P(z|z')\mathrm{d}z
\end{equation}

The Eq.\ref{eq:target_collapse} indicates that the pixel reconstruction task is minimizing the distance between representations of masked image $z'$ and original image $z$.
Now, let's zoom the lens to multiple steps, MIM can receive multiple different masked images of the original image.
By training the network to reconstruct original images from different masked ones, MIM minimizes the distance between those different masks.
Fig.~\ref{fig:MIM} shows a diagram of such training behavior, where different masking result lies at a different location of a hyperplane and construct the semantic space.
During the reconstruction training process, data points in the semantic space are pushed to align with the location of the original image.
Consequently, as they are aligned with the same data point, the distance between each data point in semantic space is also minimized.
While misalignment is a false aggregation of data points that have skewed semantics (orange dot in Fig.~\ref{fig:MIM}).
For other MIM reconstruction targets, such as visual tokens in BEiT  \cite{bao2021beit} and iBOT  \cite{zhou2021ibot}, it is easy to verify that they also share similar training behavior.
A similar conclusion has also been reported in  \cite{zhang2022mask}, however, they focus on the dimensional collapse issue in MAE and neglect the misalignment problem of MIM.
In practice, semantics are not evenly distributed in images.
These semantics are likely to be ignored by random masking strategies.
As the MIM pulls masked samples together, two images with different semantics are miss-aligned.
If the changed semantics is an important clue for image understanding, such a problem can seriously affect downstream performance.
To this end, we propose a new sampling strategy to select as representative patches as possible.

\subsection{Determinantal Point Process}
Our core technical innovation is modeling the patch masking process with DPPs.
To this end, we start with a high-level overview of DPPs. 
\vspace{0.1cm}

\noindent\textbf{Brief intro.}
A determinantal point process (DPPs) is a distribution over configurations of points. 
The defining characteristic of the DPP is that it is repulsive, which makes it useful for modeling diversity \cite{kulesza2012determinantal}.
Formally, a point process $\mathcal{P}$ on a discrete set $S = \{1, 2, ..., N\}$ is a probability measure on $2^S$, the set of all subsets of $S$.
$\mathcal{P}$ is called a determinantal point process if, when $A$ is a random subset drawn according to $\mathcal{P}$, we have,
\begin{equation}
    \label{eq:definition}
    \mathcal{P}(Y=A) \propto \operatorname{det}\left(L_A\right),
\end{equation}
where $L \in R^{N\times N}$ is a real, symmetric, positive semi-definite kernel, and $L_A \in R^{|A| \times |A|} $ is a submatrix of $L$ indexed by elements of $A$.
Note this is an unnormalized probability of sampling a set of $A$.
The normalization constant is defined as the sum of the unnormalized probabilities over all subsets of the $S$, \textit{i.e.}
$
\sum_{A \subseteq S} \operatorname{det}\left( L_A \right).
$
We can compute the normalized constant by the following theorem  \cite{kulesza2012determinantal}:
\begin{theorem}
For any $A \subseteq S$:
\begin{equation}
\sum_{A \subseteq Y \subseteq S} \operatorname{det}\left(L_Y\right)=\operatorname{det}\left(L+I_{\bar{A}}\right),
\end{equation}
where $I_{\bar{A}}$ is a diagonal matrix such that $I_{ii} = 0$ for indices $i \in A $ and $I_{ii} = 1$ for $ i \in \bar{A}$.
\end{theorem}
Setting $A = \emptyset$, we obtain the following corollary:
\begin{corollary}
\begin{equation}
\sum_{A \subseteq S} \operatorname{det}\left(L_A\right)=\operatorname{det}\left(L+I_{S}\right).
\end{equation}
\end{corollary}
Therefore, for any $A \subseteq S $, we can compute its probability by:
\begin{equation}
\label{eq:prob}
\mathcal{P}(Y=A)=\frac{\operatorname{det}\left({L}_A\right)}{\operatorname{det}\left(L+I \right)},
\end{equation}
where $I$ is the identity matrix.

\vspace{0.1cm}

We give a simple example to illustrate how DPPs model diversity. 
Suppose we have two patches $i, j$ to select, \textit{i.e.} $A = \left\{i, j\right\}$, we denote the vector of each patch as $S_i, S_j \in R^{1\times n}$, where $n$ is the dimension of elements in $S$.
We can compute the L-ensemble $L_{ij} = S_i^T S_j$ and their co-occurrence probability by Eq.\ref{eq:prob}. 
The numerator of Eq.\ref{eq:prob} can be written as:
 $det(L_A) = L_{ii} \times L_{jj} - L_{ij} \times L_{ji}$.
Note that $L_{ij}$ and $L_{ji}$ measure the similarity between elements $i$ and $j$, being more similar lowers the probability of co-occurrence. 
On the other hand, when the subset is very diverse, \textit{i.e.} $L_{ij} \times L_{ji}$ becomes small, the determinant is bigger and correspondingly its co-occurrence is more likely. 
The DPP thus naturally diversifies the selection of subsets.

Unfortunately, to our best knowledge, the implementation of exact DPP sampling needs matrix decomposition  \cite{kulesza2012determinantal, GPBV19}, which is an unacceptable computation cost during the training iteration.
Thus, to apply DPPs in MIM, an approximation is needed.
\vspace{0.1cm}

\noindent\textbf{Greedy approximation.}
Considering we only select the subset $Y$ with the highest probability under a cardinality constraint $n$, then such problem can be defined as:
\begin{equation}
    \label{map}
    Y_{\text{MAP}} = \mathop{\arg\max}\limits_{Y \subseteq S} \mathrm{det}(L_Y).
\end{equation}
This problem is known as maximum a posteriori (MAP) inference and has been proved as an NP-hard problem in DPPs  \cite{ko1995exact}.
\begin{algorithm}
\caption{Greedy DPPs Sampling for MIM}\label{alg:DPP}
\begin{algorithmic}
\Require image patches $S$, Purge ratio $\tau$, subset length $N$;
\State $S \gets \mathrm{shuffle}(S)$
\State $L \gets \mathrm{kernel}(S)$
\State $Y_g = $[]
\State $d=\mathrm{zeros}(\mathrm{len}(S))$
\While{$N \geq 0$}
    \State $d \gets \mathrm{updat}e(L, Y_g, d)$ \Comment{Follow  \cite{chen2018fast}}
    \If{$\max(d)\geq \tau$}
        \State $Y_g.\mathrm{append}(\text{argmax}(d))$
    \Else
        \State $Y_g.\mathrm{append}(\mathrm{randomSelect}(d))$
    \EndIf
    \State $N \gets N - 1$
\EndWhile
\end{algorithmic}
\end{algorithm}
Instead, the greedy algorithm is widely used for approximation \cite{nemhauser1978analysis} for MAP inference, justified by the fact that the log-probability of set in DPPs $f(Y) = {\rm log}\ det(L_Y)$ is sub-modular \cite{gillenwater2012near}.
Thus, the selection process of our method can be described as follows:
\begin{equation}
    \label{eq:selection}
    j = \mathop{\arg\max}\limits_{i\in S\setminus Y_g} f(Y_g \cup \{i\}) - f(Y_g),
\end{equation}
where $Y_g$ is the subset of $Y$. 
In each iteration, we add an item that maximizes the marginal gain to $Y_g$, until the maximal marginal gain emerges negative or goes against the cardinality constraint.
We adopt a fast implementation of the greedy MAP inference algorithm for DPPs following  \cite{chen2018fast}.
Formally, since $L$ is a PSD matrix,

the Cholesky decomposition of $L_{Y_g}$ is available as
\begin{equation}
    \label{vv}
    L_{Y_g} = V V^T,
\end{equation}
where ${\rm V}$ is an invertible lower triangular matrix. 
For any $i \in Z \setminus Y_g $, the Cholesky decomposition of $L_{Y_g \cup \{i\}}$can be derived as:
$$
L_{Y_{g} \cup\{i\}}=\left[\begin{array}{cc}
L_{Y_{g}} & L_{Y_{g}, i} \\
L_{i, Y_{g}} & L_{i i}
\end{array}\right]=\left[\begin{array}{cc}
V & {0} \\
c_i & d_i
\end{array}\right]\left[\begin{array}{cc}
V & {0} \\
c_i & d_i
\end{array}\right]^{\top},
$$
where row vector $c_i$ and scalar $d_i \geq 0$ satisfies:
\begin{equation}
\label{eq:cd}
V c_i^{\top} =L_{Y_{g}, i}, \qquad
d_i^2 =L_{i i}-\left\|c_i\right\|_2^2.
\end{equation}
Then the determinate of  $L_{Y_{g} \cup\{i\}}$ can be written as 
\begin{equation}
    \label{12}
\operatorname{det}\left(L_{Y_{g} \cup\{i\}}\right)=\operatorname{det}\left(V V^{\top}\right) \cdot d_i^2=\operatorname{det}\left(L_{Y_{g}}\right) \cdot d_i^2.
\end{equation}
Therefore, Eq.\ref{map} is equivalent to select the element $i$ with maximum ${d_i}^2$.

After solving the equation, the Cholesky factor of $L_{Y_{g}}$can therefore be efficiently updated after a new item is added to ${Y_{g}}$.
With these approximations, the selecting process can be fit in the GPU training loops.
In our experiments, the acceleration ratio is up to 10 times faster with respect to exact DPPs sampling and brings the time cost of DPPs to the same level of random, more details can be found in the supplementary.
\vspace{0.1cm}

\begin{figure}[t]
  \centering
  \begin{subfigure}{0.3\linewidth}
  \includegraphics[width=0.9\linewidth]{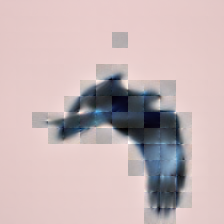}
    \caption{$\tau = 0$}
    \label{fig:sampling-a}
  \end{subfigure}
  \hfill
  \begin{subfigure}{0.3\linewidth}
  \includegraphics[width=0.9\linewidth]{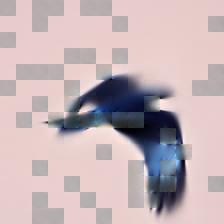}
    \caption{$\tau = 0.9$}
    \label{fig:sampling-b}
  \end{subfigure}
  \hfill
    \begin{subfigure}{0.3\linewidth}
  \includegraphics[width=0.9\linewidth]{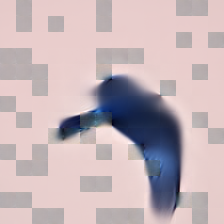}
    \caption{$\tau = 1$}
    \label{fig:sampling-c}
  \end{subfigure}
  \caption{
    Qualitative comparison of three different purge ratios, a suitable purge ratio can preserve the semantics of original images while maintaining the variety of augmented inputs.}
  \label{fig:sampling}
  \vspace{-3mm}
\end{figure}
\subsection{Purge misalignment with DPPs}
In this section, we introduce two key factors of DPPMask: kernel and purge ratio.

\noindent\textbf{Kernels.}
A common-used type of kernel is the class of Gaussian kernels \cite{tremblay2019determinantal, launay2019determinantal}.
Defined by
\begin{equation}
    \label{eq:kernel1}
    \forall S_i, S_j \in S,\qquad  L_{ij} = \mathrm{exp}(- \frac{||S_i - S_j||^2}{\epsilon}).
\end{equation}
Where $\epsilon$ is called the bandwidth or scale parameter.
This kernel depends on the squared Euclidean distance between the intensity values of pairs of patches. 
It is often used as a similarity measure on patches. 
The value of the parameter $\epsilon$ has an impact on how repulsive the DPPs are. 
However, note we only choose patches that maximize Eq.\ref{eq:selection}, and the value of $\epsilon$ influences a little to model performance.
This is because $\epsilon$ will \textit{not} change the order of distances between patches.
Thus, for better numerical stability, we normalize each patch before computing the distances and set $\epsilon$ to 1 empirically.
\vspace{0.1cm}

\noindent\textbf{Purge ratio.}
As shown in Fig.~\ref{fig:MIM}, DPPMask aims to purge those cases in that semantic information has been changed by masking.
However, DPPMask can get over-purged in some cases.
As Fig.~\ref{fig:sampling} shows, due to patches of the sky being too similar to each other, then the greedy selection will only focus on the foreground, as it is more diverse than the background.
This situation makes the MIM task too easy and purges most of useful augmented inputs, which is not helpful in feature learning.
A simple modification can tackle this problem.
Instead of letting the selection process hit the cardinality constraint, we set a parameter called purge ratio $\tau \in (0, 1)$  as the threshold of maximal marginal gain.
Concretely, in each iteration, we monitor the distance of the next patch to the selected subsets, if the distance is below the purge ratio $\tau$, abort the greedy selection process and fill the subset will random patches.
The purge ratio plays a role of adjust how "severe" the DPPMask is.
In particular, $\tau = 0$ indicates the selection process becomes fully greedy and $\tau = 1$ indicates fully random sampling.
Fig.~\ref{fig:sampling} shows greedy selection under three different purge ratios, a higher purge ratio can prevent DPPMask get over-purged and maintain the input diversity for training the network.

In this section, we first analyze the training behavior of MIM, then we give our method namely DPPMask which uses DPPs to model the repulsion of image patches, in order to sample the most representative patches and preserve the original semantic information of images.
We summarize our algorithm in Alg.~\ref{alg:DPP}.

\section{Experiment}
\subsection{Implementation details}
To examine the effectiveness of our method, we perform DPPMask on two representative MIM methods: MAE \cite{he2022masked}, iBOT \cite{zhou2021ibot}, which represent two different MIM frameworks: pixel reconstruction and feature contrast.
For MAE, images are patched by convolutional kernels and added with a position embedding, after that, we compute the distances of patches.
For iBOT, images are fed into a teacher model to get semantic tokens, which we compute distances based on.
Compare to direct computing with pixel intensities, semantic tokens may contain more useful information for partitioning images.
For example, to identify instances that share similar appearances.

We adopt two different scales of backbones, ViT-Base and ViT-Small for MAE and iBOT respectively. 
We mainly evaluate our algorithms on the ImageNet-100 dataset, which is derived from ImageNet-1K \cite{russakovsky2015imagenet, shi2022adversarial}.
We train MAE and iBOT for 400 and 100 epochs respectively.
Unfortunately, our computational resources can not support us to make out larger-scale experiments, such as more training epochs and heavier backbones. We leave this for future work.
We set the masking ratio is set to $0.75$ for MAE and $0.7$ with $0.05$ variance for iBOT by default.

\begin{table}
  \centering
    \scalebox{1.0}{
  \begin{tabular}{c c c c}
    \toprule
    Method & Pre-train loss & Linear prob. & Fine-tuned\\
    \midrule
    $\tau=0.6$ & 0.417 & - & 89.67 \\
    $\tau=0.8$ & 0.434 & 62.58 &\textbf{89.67} \\
    $\tau=0.9$ & 0.440 & 63.22 & 89.56\\
    \midrule
    MAE & 0.444 & \textbf{67.08} & 89.45 \\
    \bottomrule
  \end{tabular}}
  \caption{Detailed results of MAE+DPPMask on ImageNet-100. The best of each metric are marked in bold.}
  \label{tab:tau-mae}
  \vspace{-3mm}
\end{table}

\begin{table}
  \centering
  \scalebox{0.93}{
  \begin{tabular}{c c c c c}
    \toprule
    Method & NMI & ACC & Linear prob. & Fine-tune \\
    \midrule
    $\tau=0.6$ & 0.518 & 67.88 & 72.84 & 87.58 \\
    $\tau=0.8$ & 0.522 & 68.04 & 73.56 & 87.64 \\
    $\tau=0.9$ & \textbf{0.525} & \textbf{68.68} & \textbf{73.60} & \textbf{87.84}\\
    \midrule
    iBOT & 0.522 & 68.28 & 73.30 & 87.44\\
    iBOT+AttMask & 0.512 & 67.32 & 72.30 & 87.44 \\
    \bottomrule
  \end{tabular}}
  \caption{Detailed results of iBOT+DPPMask on ImageNet-100. The best of each metric are marked in bold.}
  \label{tab:tau-ibot}
  \vspace{-3mm}
\end{table}

\subsection{A detailed study of misalignment}
We give our main result in Tab.~\ref{tab:tau-mae} and Tab.~\ref{tab:tau-ibot}.
The final feature vector for classification is obtained by global pooling.
For fine-tuning tasks, we run each setting with three random seeds and report their average performance.
We train the iBOT model under the fine-tuning and linear probing parameter setting of MAE for reducing the experiment's complexity.
\vspace{0.1cm}

\noindent\textbf{The performance gain brings by DPPMask.}
We first observed a steady performance gain on fine-tuning tasks in both MAE and iBOT frameworks.
In MAE, we make $0.2\%$ accuracy gains.
In iBOT, we make $0.4\%$ accuracy gains. 
This shows our method can improve the representation power by purging the misalignment samples.
As the threshold $\tau$ increases, the pre-train loss becomes smaller in response.
\begin{figure}[t]
  \centering
  \includegraphics[width=0.9\linewidth]{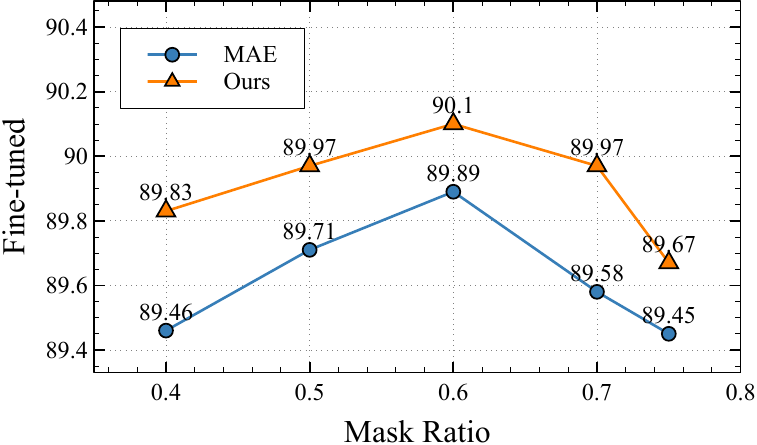}
  \caption{Validation accuracy of MAE on ImageNet-100 under different masking ratios.}
  \label{fig:maskratio}
  \vspace{-3mm}
\end{figure}
\vspace{0.1cm}
\begin{table}
  \centering
    \scalebox{1.0}{
  \begin{tabular}{c c c c}
    \toprule
    Method & Backbone & Epochs & Fine-tuned\\
    \midrule
    MAE & ViT-Base & 400 & 82.9 \\
    MAE+AMT & ViT-Base & 400 & 82.8 \\
    SemMAE & ViT-Base & 800 & 83.3 \\
    \midrule
    MAE+DPPMask & ViT-Base & 400 & \textbf{83.3}\\
    \bottomrule
  \end{tabular}}
  \caption{Comparison with other sampling methods on ImageNet-1K. The best of each metric are marked in bold.}
  \label{tab:compare}
  \vspace{-3mm}
\end{table}
However, when $\tau$ is too low, greedy sampling goes over-purged and makes the task too simple for the network to learn useful features.
Besides, iBOT and MAE show different preferences of $\tau$, this is because we apply DPPs on the output of the teacher model, which is a different distribution from MAE.
For linear probing, we notice a significant performance drop of MAE.
This does not surprise us as we analyze in Sec.~\ref{sec:ci}.
DPPs make the sample space shrunk in order to purge the misalignment samples. 
Aggregating fewer samples together makes the feature space more continuous and less linear separable.
Notably, the features also become more precise to describe an image, which can reflect the fine-tuned performance.
This behavior is not observed on iBOT, which uses extensive augmentation to further expand the scale of positive samples.
Instead, iBOT got $0.3\%$ performance gain on linear probing as well as cluster performance (NMI and ACC), which proves our method indeed purged improper samples that hurt feature learning.
Tab.~\ref{tab:compare} shows our method alongside other advanced sampling methods
on ImageNet-1K, our methods surpassed other sampling methods.

\noindent\textbf{DPPMask makes the MIM task more reasonable.}
Another key factor of successfully applying MIM is the masking ratio of input images \cite{he2022masked, xie2022simmim}.
It should be high enough to construct a meaningful reconstruction target while preventing the task from degenerating to simply copying-pasting from neighboring patches.
However, the root cause of the misalignment problem is also the aggressive masking strategy.
To better understand the relationship between the masking ratio and the misalignment problem, we study the fine-tuned performance of MAE under different masking ratios.
\begin{figure}[t]
  \centering
  \begin{subfigure}{0.18\linewidth}
  \includegraphics[width=0.9\linewidth]{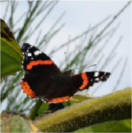}
    \caption{Orig.}
    \label{fig:bg-ori}
  \end{subfigure}
  \hfill
  \begin{subfigure}{0.18\linewidth}
  \includegraphics[width=0.9\linewidth]{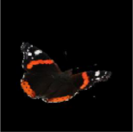}
    \caption{O.F.}
    \label{fig:bg-of}
  \end{subfigure}
  \hfill
    \begin{subfigure}{0.18\linewidth}
  \includegraphics[width=0.9\linewidth]{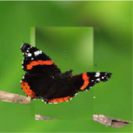}
    \caption{M.S.}
    \label{fig:bg-ms}
  \end{subfigure}
    \hfill
    \begin{subfigure}{0.18\linewidth}
  \includegraphics[width=0.9\linewidth]{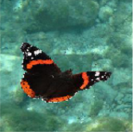}
    \caption{M.R.}
    \label{fig:bg-mr}
  \end{subfigure}
    \hfill
    \begin{subfigure}{0.18\linewidth}
  \includegraphics[width=0.9\linewidth]{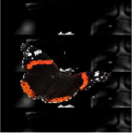}
    \caption{M.N.}
    \label{fig:bg-mn}
  \end{subfigure}
  \caption{
    Examples of background challenge.}
  \label{fig:bg-challenge}
  \vspace{-3mm}
\end{figure}
\begin{table}
  \centering
  \scalebox{0.9}{
  \begin{tabular}{l c c c c c}
    \toprule
    Method & Orig. & O.F. & M.S. & M.R. & M.N \\
    \midrule
    iBOT &  56.05& 42.32& 48.07& 39.70&37.23 \\
    iBOT+AttMask &54.22& 41.01& 46.15&37.98& 36.02\\
    iBOT+DPPMask &\textbf{56.47} & \textbf{43.36}&\textbf{49.65} &\textbf{40.59} &\textbf{38.44} \\
    \midrule
    MAE & 55.46 & 43.04 & 47.36 & 40.12 & 38.30 \\
    MAE+DPPMask & \textbf{56.32} & \textbf{44.12} & \textbf{47.73} & \textbf{40.74} & \textbf{39.43} \\
    \bottomrule
  \end{tabular}}
  \caption{Performance on background challenge.}
  \label{tab:bg_challenge}
  \vspace{-3mm}
\end{table}
We perform each experiment three times and report their mean accuracy of ImageNet-100.
For DPPs sampling, we run two values of $\tau$, $0.90$ and $0.85$, we report the higher performance.
As Fig.~\ref{fig:maskratio} shows, we find that the original $0.75$ masking ratio is not the optimal setting for fine-tuned performance.
Instead, lowering the masking ratio significantly improves the accuracy, this is further evidence of the impact of the misalignment problem on feature learning, as a higher masking ratio raises the probability of misalignment.
With DPPs sampling equipped, our method has achieved higher performance in all masking ratio settings.
In the masking ratio $0.7$, the maximal performance boost reached $0.4\%$.
When the masking ratio gets lower, the pretext task actually becomes more simple, which leads to a performance drop.
Notably, we make a better result than the best in MAE (masking ratio at 0.6) with both higher and lower masking ratios ($0.5$, $0.7$).
This is meaningful, as the reconstruction problem becomes more simple while our method still performs better than MAE, which shows our sampling method makes the pre-train task more reasonable rather more simple. 
\subsection{Robustness}
The misalignment problem makes the network align images with different semantics.
In some severe cases, the network may are required to align the original image with the background. 
Thus, the misalignment problem can interfere with the network decision by letting the network more focusing the background of images.
To verify this, we evaluate the quality of the learned feature on the background challenge \cite{xiao2020noise}.
We run fine-tuned models of each method on 4 different variations from the original image.
Each variation replaces the original background with empty (O.F.), with another image in the same class (M.S.), with a random image in any class (M.R.), or with an image from the next class (M.N.), examples are shown in Fig.~\ref{fig:bg-challenge}. Tab.~\ref{tab:bg_challenge} shows the performance of our method on the background challenge.
\begin{table}
  \centering
\scalebox{1.0}{
  \begin{tabular}{l c c c}
    \toprule
    Method & mAP & F$1_{all}$ & F$1_{class}$ \\
    \midrule
    iBOT & 63.16 & 64.94 & 55.78 \\
    iBOT+AttMask & 63.26 & 65.31 & 56.36 \\
    iBOT+DPPMask & \textbf{63.78} & \textbf{65.72} & \textbf{56.40} \\
    \midrule
    MAE & 68.95 & 69.05 & 61.24 \\
    MAE+DPPMask & \textbf{69.56} & \textbf{69.59} & \textbf{61.86} \\
    \bottomrule
  \end{tabular}}
  \caption{Multi-label classification accuracy on COCO.}
  \label{tab:coco}
  \vspace{-3mm}
\end{table}
\begin{table}
  \centering
\scalebox{0.9}{
  \begin{tabular}{l c c c c}
    \toprule
    Method & ACC & F$1_{macro}$ & F$1_{micro}$ &F$1_{weighted}$ \\
    \midrule
    MAE & 80.69 & 44.81 & 45.45 & 46.46 \\
    MAE+DPPMask & \textbf{80.85} & \textbf{45.08} & \textbf{45.85} & \textbf{47.24} \\
    \bottomrule
  \end{tabular}}
  \caption{Multi-label classification accuracy on CLEVR.}
  \label{tab:clevr}
    \vspace{-3mm}
\end{table}
\begin{figure*}[t]
  \centering
  \includegraphics[width=1\linewidth]{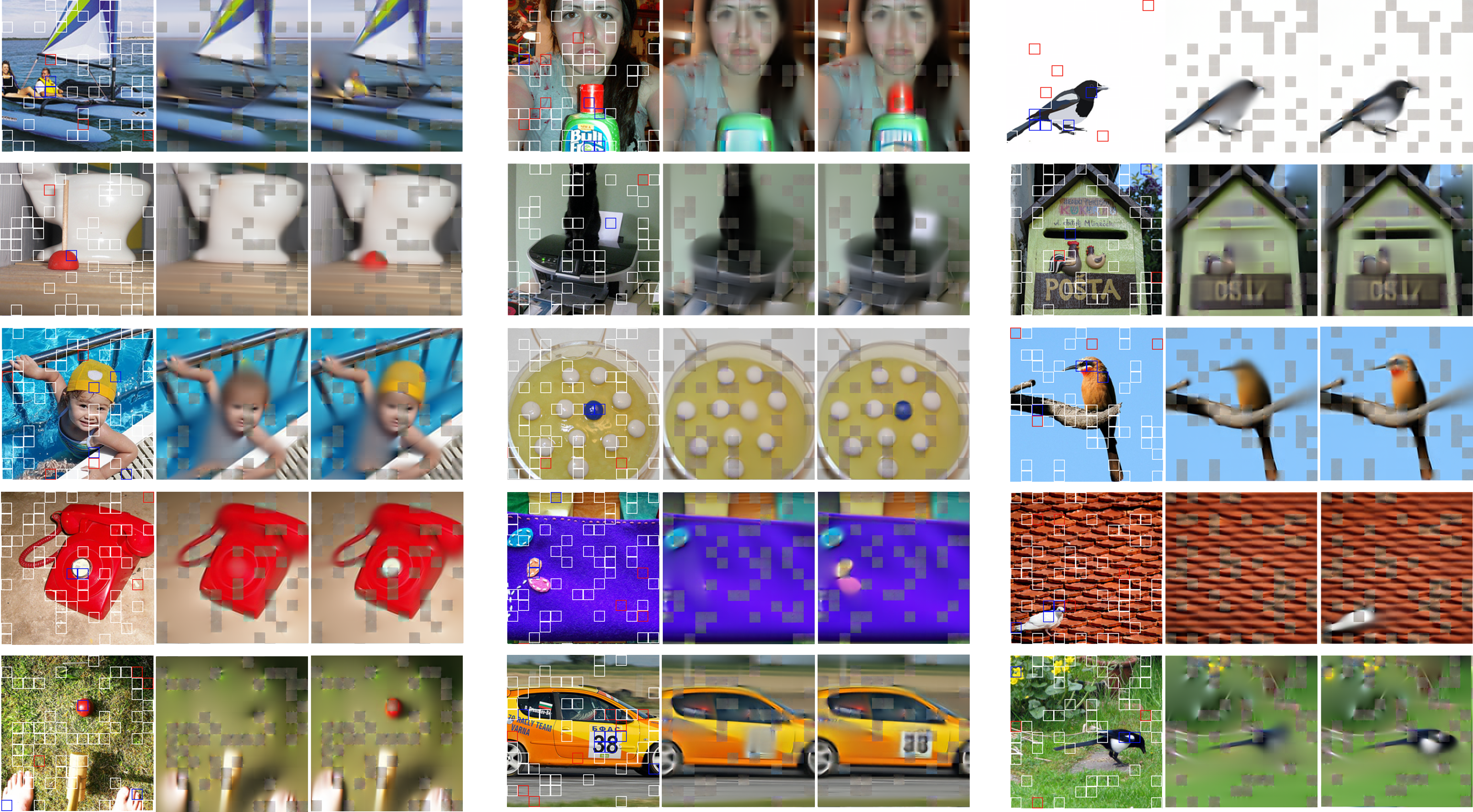}
    \caption{Comparison of DPPs sampling and random sampling, each triplet indicates the original image (right), reconstruction result with random sampling (middle), and DPPs sampling (left). The white boxes represent patches selected by both random and DPPs, while red boxes are for random sampling and blue boxes are for DPPs sampling. The threshold $\tau$ is set to 0.8, best views with zoom-in.}
    \label{fig:quali}
  \vspace{-3mm}
\end{figure*}

Both iBOT and MAE witnessed a steady performance gain in four different variations of original images.
Notably, our method on MAE achieves $1.08\%$ and $1.13\%$ improvements on O.F. and M.N images respectively, which is higher than the original images ($0.86\%$). 
Such results were also observed in the iBOT framework.
Our method achieves $0.42\%$ improvements on original images, while other variations both improved by a large margin.
In particular, we achieve $1.58\%$ improvements on M.S. which largely surpasses the original improvements.
This shows our methods are more robust to background changes, as we do not impose the network to align the background to the original image as the random strategy does.

\subsection{Multi-label classification}
To further examine the influence of the misalignment problem, we also test our method on a multi-label classification task.
An intuitive understanding is that: the network can not reflect the semantic changes, which are trained to reconstruct objects whether are been masked or not.
Where in multi-label classification, every semantics is important, which makes them suitable to test whether the network is capable to extract whole information of images.
We test our method on CLEVR \cite{johnson2017clevr} and MS-COCO \cite{lin2014microsoft} datasets, which are widely used in multi-label classification.
The CLEVR dataset contains 24 binary labels, each indicating the presence of a particular color and shape (8 × 3) combination in the image \cite{shi2022adversarial}.
For the MS-COCO dataset, we report the following fine-tuned performance on the validation set: mean average precision (mAP), average per-class F1 score (F$1_{class}$), and the average overall F1 score (F$1_{all}$)  \cite{ridnik2021asymmetric}.
For the CLEVR dataset, we train MAE with and without DPPs for 200 epochs, we report the linear probing and fine-tuned performance of F$1_{micro}$, F$1_{macro}$ and F$1_{weighted}$, where ‘micro’ evaluate F1 across the entire dataset, ‘macro’ evaluates an unweighted average of per-label F1 score, and ‘weighted’ scale the per-label F1 score by a number of examples when taking the average.
We show our result in Tab.~\ref{tab:coco} and Tab.~\ref{tab:clevr}.
Our method shows better performance on both COCO and CLEVR datasets.
This shows our sampling method can select more informative patches for the network to reconstruct or align, which reduces the impact of the misalignment problem.

\subsection{Qualitative analysis of DPPs sampling}
To better understand the sampling behavior of DPPs, we compare the DPPs sampling result with random sampling.
Fig.~\ref{fig:quali} shows the MAE reconstruction result from the ImageNet validation set, each triplet from left to right indicates the original image, reconstruction result with random sampling and DPPs sampling.
For each image, we fix the random seed in order to find the difference between DPPs and random.
We show coincide patches with white boxes, patches in random sampling while not in DPPs are shown in red boxes, and patches in DPPs while not in random are shown in blue boxes.
Our experiment shows the reconstruction result of DPP sampling is better than random sampling, which proves that DPPs can represent more complete semantics than random.
In particular, the sampling result shows two important properties of DPPs.
First, DPPs can catch the appearance of each object more precisely, which is an important clue for image understanding.
For example, the slot of the mailbox is crucial evidence to classify with cabin.
Another important property is DPPs can retain more small foreground information, which is highly likely omitted in random sampling.
Such properties show our method successfully alleviates the impact of misalignment problem, and achieve better performance in feature learning.

In our experiments, the MAE does not reconstruct the unobserved semantics, indicating that false positive samples are not perfectly aligned.
A proper guess of such a phenomenon can be the diversity of ImageNet or the representative capabilities of networks.
However, despite the network does not fall into over-fit, the incorrect gradient of misalignment will still interfere with the learning process.
Our experiment also shows the potential of MIM with fewer misalignment problems.

\section{Conclusion}
In this paper, we show that uniformly random masking widely used in previous works unavoidably loses some key objects and changes original semantic information, resulting in a misalignment problem and hurting the representative learning
eventually.
To this end, we propose a new masking strategy namely the DPPMask to reduce the semantic change of the image after
masking.
We show that DPPMask can make the MIM task more reasonable by purging the misalignment of training pairs.
We hope our work can provide insights to help design a better MIM algorithm.

{\small
\bibliographystyle{ieee_fullname}
\bibliography{egbib}
}
\clearpage
\appendix
\section{Implementation details of DPPMask}
Suppose we add $i$ into the subset $Y_g \cup\{j\}$. 
From Eq.~\ref{eq:cd}, we have
\begin{equation}
    \label{eq:decomp_j}
    \left[
    \begin{array}{cc}
        V & 0 \\ 
        c_j & d_j
    \end{array}
    \right] 
    c_i^{\prime \top} = L_{Y_{g} \cup\{j\}, i}=
    \left[
    \begin{array}{c}
    L_{Y_{g}, i} \\
    L_{j i}
    \end{array}
    \right],
\end{equation}
where
\begin{equation}
\label{eq:update_c}
c_i^{\prime}=\left[
    \begin{array}{cc}
        c_i & \left(L_{j i}-\left\langle c_j, c_i\right\rangle\right) / d_j
    \end{array}\right] \doteq\left[\begin{array}{ll}c_i & e_i\end{array}\right].
\end{equation}
For updating $d_i$, we have
\begin{equation}
    \label{eq:update_d}
    \begin{aligned}
    d_i^{\prime 2}&=L_{i i}-\left\|c_i^{\prime}\right\|_2^2 \\
    &=L_{i i}-\left\|c_i\right\|_2^2-e_i^2 \\
    &=d_i^2-e_i^2.
    \end{aligned}
\end{equation}
With Eq.~\ref{eq:update_c} and Eq.~\ref{eq:update_d}, we can update $d$ incrementally.


\section{ImageNet-100 setting}
We follow the original MAE \cite{he2022masked} experiment setting, except for the learning rate of fine-tuning task.
We list our fine-tuning parameters in Tab.~\ref{tab:ft_param}.
For iBOT \cite{zhou2021ibot}, we train a ViT-small backbone with 100 epochs. 
We change the block mask strategy with random, and set masking ratio to $70\%$ with $5\%$ variation.
We list our fine-tuning parameters in Tab.~\ref{tab:ibot_param}
\begin{table}
    \centering
    \small
    \caption{Fine-tuning setting.}
\begin{tabular}{l|l} 
\label{tab:ft_param}
config & value \\
\hline 
optimizer & AdamW \cite{loshchilov2019decoupled} \\
base learning rate & $5 \mathrm{e}-4$ \\
weight decay & $0.05$ \\
optimizer momentum & $\beta_1, \beta_2=0.9,0.999$ \cite{chen2020rewon}\\
layer-wise lr decay \cite{clark2020electra,bao2021beit}& $0.65$ \\
batch size & 1024 \\
learning rate schedule & cosine decay \cite{loshchilov2016sgdr} \\
warmup epochs \cite{goyal2017accurate} & 5 \\
training epochs & $100$ \\
cutmix\cite{yun2019cutmix} & $1.0$ \\
drop path & $0.1$ \\
mixup \cite{zhang2017mixup} & $0.8$ \\
weight decay & $0.05$ \\
label smoothing \cite{szegedy2016rethinking} & $0.1$ \\
augmentation & $\operatorname{RandAug}(9,0.5)$\cite{cubuk2020randaugment}
\end{tabular}
\end{table}
\begin{table}
    \centering
    \small
    \caption{iBOT pre-train setting.}
\begin{tabular}{l|l} 
\label{tab:ibot_param}
config & value \\
\hline 
learning rate & $5 \mathrm{e}-4$ \\
teacher momentum \cite{bao2021beit} & $0.996$ \\
teacher temp & $0.07$ \\
warmup teacher temp epochs \cite{bao2021beit} & $30$ \\
out dim & $8192$ \\ 
local crops number & $10$ \\
global crops scale & $[0.25, \ 1]$ \\
local crops scale & $[0.05, \ 0.25]$ \\
mask ratio & $0.7$ \\ 
mask ratio var & $0.05$ \\
masking prob & $0.5$ \\
\end{tabular}
\end{table}

\begin{table}[t]
\centering
\caption{Training efficiency of DPPMask}
\label{tab:speed}
    \resizebox{0.48\textwidth}{!}{
    \large
    \begin{tabular}{*{4}{c}}
        \toprule
       Method & DPP &  DPP (Greedy approximation) & Random \\
        \midrule
        Training speed & 2.7442s/it  & 0.2609 s/it & 0.2129 s/it \\
        \bottomrule
    \end{tabular}
    }
    \vspace{-1em}
\end{table}

\section{Greedy approximation performance}
We examine the approximation performance with respect to original DPPs.
We run each setting with one entire epoch and report the mean time cost of each iteration.
As Tab.~\ref{tab:speed} shows, the greedy approximation achieved 10x faster than original DPPs, which makes it possible to fit into a GPU training loop.

\section{More discussion about relative works}
Our method is different from recent masking strategies.
Recent strategies can roughly divide into two lines of work, learning-based and attention-based.

\noindent\textbf{Learning-based strategies} include ADIOS\cite{shi2022adversarial} and SemMAE\cite{li2022semmae}.
They both need extra learning parameters.
ADIOS trains a network to propose masks adversarially, in order to find out more meaningful masks for MIM tasks.
However, the semantic meaningful masks proposed by the network are hard to predict, which is less explainable.
SemMAE separates the mask learning process from pre-training and makes the training process into two stages.
The type of masks they learned are more like semantic parts, such as heads, arms, etc.
However, as the semantics varies in images, the number of classes is hard to define, therefore weakening the application of such methods.

\noindent\textbf{Attention-based strategies} include AttMask\cite{kakogeorgiou2022hide} and AMT\cite{gui2022good}. 
This line of work selects patches according to the attention map.
Despite different policies to manipulate the attention map, both intend to retain some patches with high attention scores to give a "hint" to the model as such patches are more likely to have more semantics.
Note this policy aligns with our ideas.
However, they do not associate it with misalignment problems in MIM, thus leading to an inferior policy.
Furthermore, attention maps require an extra forward pass to compute, which brings more computation. 

\section{Alignment versus diversity}
\begin{figure}
  \centering
  \begin{subfigure}{0.45\linewidth}
  \includegraphics[width=0.9\linewidth]{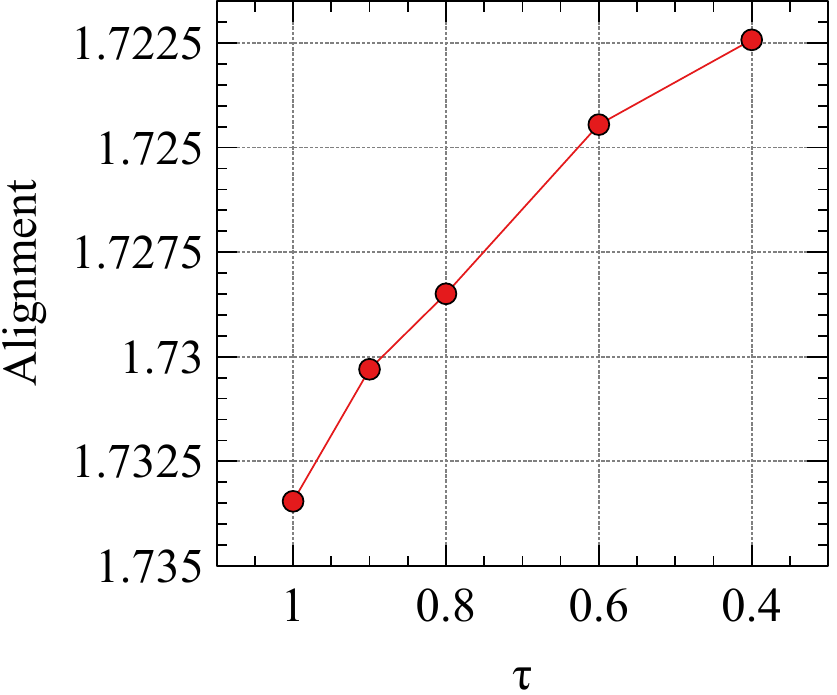}
    \caption{Alignment}
    \label{fig:align}
  \end{subfigure}
  \hfill
  \begin{subfigure}{0.45\linewidth}
  \includegraphics[width=0.9\linewidth]{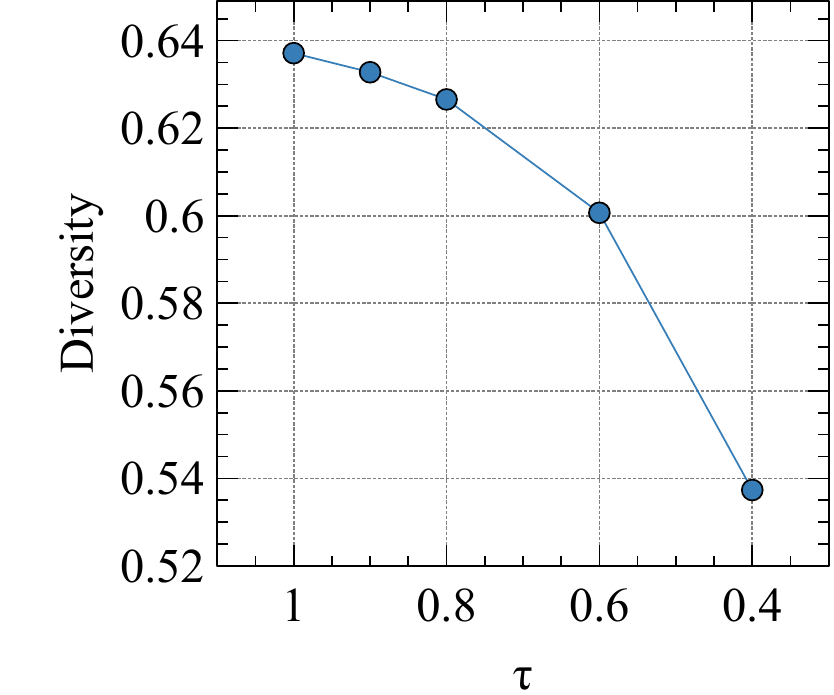}
    \caption{Diversity}
    \label{fig:div}
    \end{subfigure}
    \caption{Alignment and diversity of different $\tau$s, $\tau = 1$ indicate random masking.}
    \label{fig:aligndiv}
\end{figure}

As we discussed in \ref{sec:ci}, DPPMask aims to purge the training pairs that are polluted by misalignment problems.
However, the network also needs irrelevant information between different masks to perform feature learning which can be measured as the variance of sampled masks.
Here, we show that DPPMask achieves the adjustment between alignment and diversity. 
We first obtain the original semantic representation by feeding the network with unmasked images and saving the \textit{cls} token.
Then, we obtain the masked semantic representation by saving the \textit{cls} tokens of masked images under different masking strategies.
We compute the L2 distance between masked semantics and original semantics to illustrate the alignment of different masking strategies.
For masked semantic representation, we run 5 independent trails and compute the L2 distance between each trail to illustrate the diversity of different masking strategies.
As shown in \ref{fig:aligndiv}, random masking has the most diversity, but it also suffers from the misalignment problem, i.e. the farthest distance between masked semantics and original semantics~\ref{fig:align}.
As the $\tau$ decrease, the distance between masked semantics and original semantics has been reduced, indicating more alignment to the original semantics.
However, a lower $\tau$ cause less diversity of different masks, which can purge some useful training pairs and is not helpful for feature learning~\ref{fig:div}.

\section{Broader impact}
Despite the eye-catching performance of MIM algorithms, what makes a good mask for MIM tasks still remains unclear.
DPPMask provides a possible answer to this question.
By analog to the \textit{InfoMin} principle of contrastive learning.
We conclude two properties of MIM.
Masked images should retain the original semantics while minimizing shared information from different masks. 
Minimizing shared information can be achieved by setting a high mask ratio, while how to retain the original semantics is a non-trivial problem.
Furthermore, DPPMask also models the probability-of-co-occurrence of each patch and thus can serve as a potential tool to study the relationship between such two properties.
\clearpage
\begin{figure*}[t]
  \centering
  \includegraphics[width=1\linewidth]{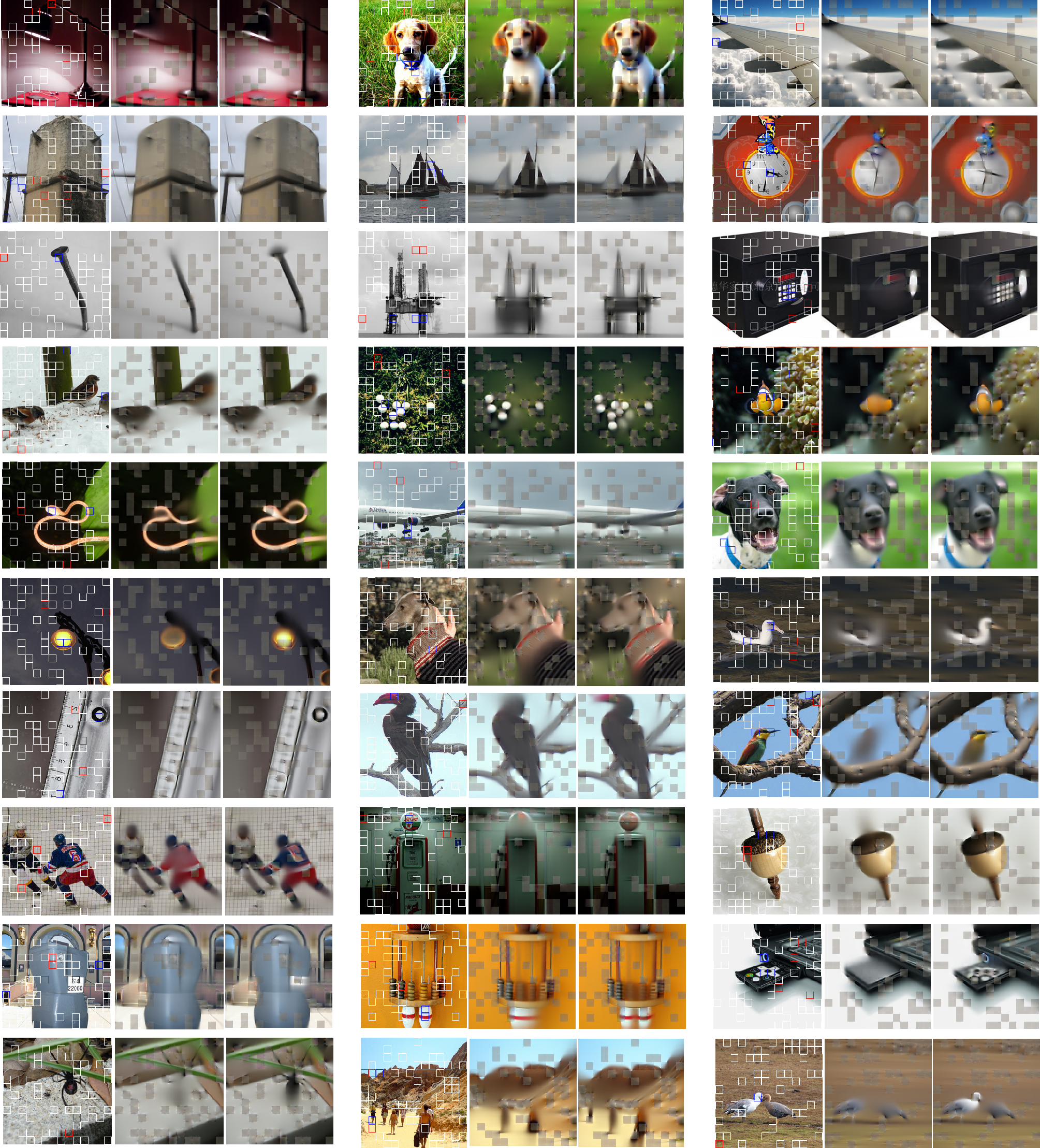}
    \caption{More qualitative samples. Each triplet indicates the original image (right), reconstruction result with random sampling (middle), and DPPs sampling (left).}
    \label{fig:quali}
\end{figure*}
\end{document}